\documentclass[10pt,twocolumn,letterpaper]{article}
\usepackage[pagenumbers]{cvpr}

\usepackage{iftex}
\ifPDFTeX\else
  \usepackage{fontspec}
\fi
\usepackage{tikz-cd}
\usepackage{multirow}
\definecolor{cvprblue}{rgb}{0.21,0.49,0.74}
\usepackage[pagebackref,breaklinks,colorlinks,allcolors=cvprblue]{hyperref}
\AtEndPreamble{\crefname{appendix}{Supp.~Sec.}{Supp.~Secs.}\Crefname{appendix}{Supplementary Section}{Supplementary Sections}%
  \crefname{figure}{Figure}{Figures}\Crefname{figure}{Figure}{Figures}%
  \crefname{table}{Table}{Tables}\Crefname{table}{Table}{Tables}}

\def\confName{CVPR}
\def\confYear{2027}
\newif\ifrealworld
\realworldfalse

\title{DexFormer 2.0: Towards a Cross-Embodiment\\
Foundation Policy for Dexterous Grasping}
\author{Lixin Xu$^{1,2,*}$\quad Ke Zhang$^{1,2,*}$\quad Xinyu Liu$^{2,*}$\quad Jie Tan$^{2,3}$\quad Senming Tan$^{2}$\\
Jiaju Yin$^{1,2}$\quad Yating Feng$^{1,2}$\quad Xiaoyi Lin$^{4}$\quad Xiyue Dong$^{1,2}$\quad Renjing Xu$^{1,\dagger}$\\[4pt]
{\normalsize $^1$HKUST(GZ)\quad $^2$BrainCo\quad
$^3$Chongqing University\quad $^4$Wuhan University}\\[2pt]
{\small $^*$Equal contribution\quad $^\dagger$Corresponding author}\\[3pt]
{\small Project website: \url{https://davidlxu.github.io/dexformer2-web/}}}

\makeatletter
\appto\@maketitle{%
  \begin{center}
    \includegraphics[width=\textwidth]{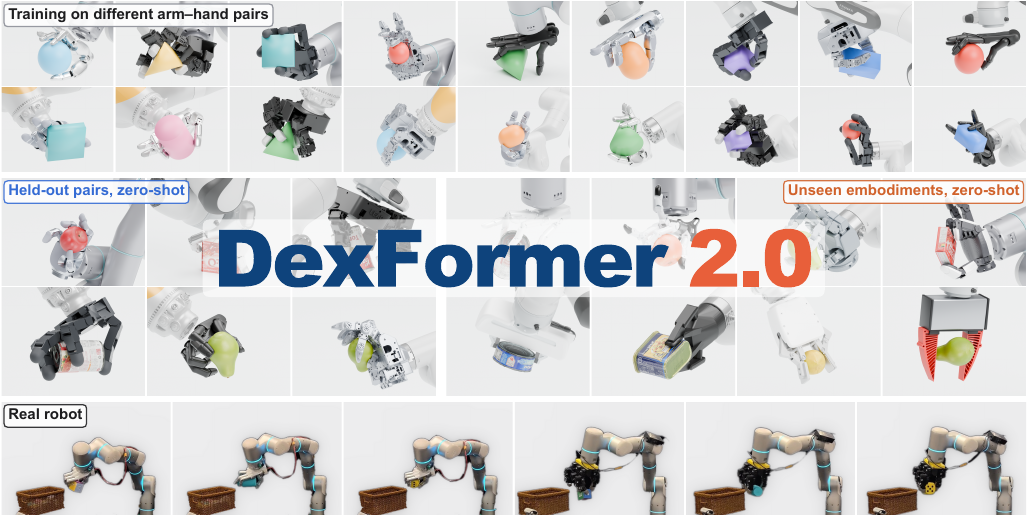}
    \captionsetup{type=figure,skip=3pt}
    \caption{\textbf{DexFormer 2.0.} \rev{One policy grasps with diverse
    arm--hand embodiments. Top: the 18 training pairs on training primitives.
    Middle: \heldterm{held-out arm--hand pairs} (left) and new hands and grippers (right),
    zero-shot on YCB objects; grippers use virtual-hand retargeting. Bottom:
    distilled students on a real Flexiv arm with Revo3 and LEAP hands.}}
    \label{fig:teaser}
  \end{center}
  \vspace{4pt}
}
\makeatother

\usepackage{float}
\newcommand{\cut}[1]{#1}
\newcommand{\tocheck}[1]{#1}
\newcommand{\rev}[1]{#1}
\newcommand{\seedmark}[1]{#1}

\newcommand{\revtwo}[1]{#1}
\newcommand{\heldterm}[1]{#1}
\newcommand{\memterm}[1]{#1}
\newcommand{\sharedmark}[1]{#1}

\begin{document}
\maketitle

\begin{abstract}
\rev{Dexterous grasping policies are usually trained for one arm paired
with one hand. Every new pairing then needs its own policy, and the number
of policies grows with every arm and hand added. We ask whether a single
policy can control many arm--hand combinations and transfer to pairings
and hands it has never seen. This is hard because robots differ in the
number, layout, and dynamics of their joints, so their observations and
actions share no common interface. We present DexFormer 2.0, a single
reinforcement learning policy for cross-embodiment dexterous grasping. It
maps every robot to a canonical joint interface with validity masks,
encodes each joint's morphology with attention biased by the kinematic
tree, and uses a Transformer-XL \memterm{memory} of recent interaction to capture the
dynamics that these static descriptors do not specify. Trained on 18 combinations of four arms
and six hands, it reaches 89\% stable success on seen combinations and 51\% on six \heldterm{held-out arm--hand pairs}, outperforming GRU and DexFormer~1.0 baselines. Without
fine-tuning, it also grasps with \rev{four} new hand types and, through
retargeting, with parallel grippers. Ablations show that \memterm{memory}, the
kinematic-tree bias, and broader pairing coverage each matter for
transfer. On a physical Flexiv arm, distilled students transfer zero-shot
from simulation with two hands, including a \heldterm{held-out pair}.}

\end{abstract}

\section{Introduction}
\label{sec:introduction}

Dexterous grasping policies are commonly developed for a fixed robotic
embodiment: one arm, one hand, and one carefully matched action space.
This specialization makes policy development expensive as hardware
diversifies. Replacing either the arm or the hand changes the kinematic
chain, reachable workspace, joint limits, control response, and contact
geometry; replacing both compounds these changes. A straightforward
solution is to train a separate policy for every arm--hand pairing, but
the number of policies grows with the Cartesian product of available
arms and hands, preventing knowledge sharing across related embodiments.

Learning one policy across heterogeneous arm--hand systems is challenging
for three reasons. First, different hands expose different numbers and
arrangements of active joints, so their observations and actions do not
naturally share a fixed interface. Second, the same action value has a
different physical meaning under different joint axes, limits, link
placements, and kinematic depths. Third, morphology alone is insufficient:
closed-loop grasping also depends on contacts and embodiment-specific
dynamics that must be inferred over time.

We present \textbf{DexFormer 2.0}, a morphology-aware recurrent policy for
cross-embodiment dexterous grasping (\cref{fig:teaser}). We represent every robot through a
canonical set of 29 joint slots---seven arm joints and up to 22 hand
joints---together with validity masks. Each joint receives a 16-dimensional
morphology descriptor encoding its axis, attachment position, joint and
velocity limits, semantic role, and kinematic depth. Our MorphNode encoder
adds an attention residual biased by the robot's kinematic tree and joint
roles, and a Transformer-XL maintains a 16-step \memterm{memory} over the
fused features. A shared actor head then produces a masked Gaussian
policy over the canonical action space, using the same network across all
arm--hand pairings.

We study a single policy over four robot arms and six dexterous hands,
forming 24 arm--hand combinations. We randomly select 18 combinations for
training and reserve the remaining six for testing. Evaluation on both training and \heldterm{held-out arm--hand pairs} uses the same ten object identities, isolating
embodiment generalization from object-identity generalization. The policy
reaches 89.01\% stable success on the seen combinations and 51.18\% on the \heldterm{held-out pairs}, and it keeps 48.85\% on \heldterm{held-out pairs} when the
objects are ten unseen YCB objects~\cite{calli2015ycb}. Without
fine-tuning, it also grasps with four hand types that never appear in
training, and its actions can be retargeted to four parallel grippers.
Our contributions are:
\begin{itemize}
    \item We introduce a \textbf{joint cross-embodiment training paradigm}:
    one policy is trained on many arm--hand embodiments together, through
    canonical joint slots and validity masks. The policy, without
    fine-tuning, transfers to \heldterm{held-out pairs} and to hand
    types absent from training, and even to parallel grippers.

    \item We develop a \textbf{topology-aware recurrent policy architecture}
    that combines per-joint physical descriptors, a gated spatial-attention
    residual, Transformer-XL \memterm{memory}, and a shared actor head.
    The architecture conditions control on morphology without using a
    dedicated policy or output head for each embodiment.

    \item We show that \textbf{\memterm{temporal memory}, kinematic-tree bias, and
    pairing coverage} each matter for transfer: removing \memterm{memory} or the
    tree bias roughly halves success on \heldterm{held-out pairs}, and training
    on 18 instead of 12 combinations raises it from 5.65\% to 51.18\% at
    a matched interaction budget.
\end{itemize}

\section{Related Work}
\label{sec:related-work}

\paragraph{Dexterous grasping and sim-to-real transfer.}
Reinforcement learning has enabled contact-rich dexterous control.
Andrychowicz et al.~\cite{andrychowicz2020learning} showed that domain
randomization~\cite{tobin2017domain} can transfer simulated in-hand
manipulation policies to a physical hand, and later work extended this
route to faster reorientation and rotation~\cite{handa2023dextreme,qi2022inhand}.
For grasping, UniDexGrasp~\cite{xu2023unidexgrasp} and
UniDexGrasp++~\cite{wan2023unidexgrasp} learn object-general policies
through goal conditioning, geometry-aware curricula, and
iterative generalist--specialist learning, while
UniGraspTransformer~\cite{wang2025unigrasptransformer} distills
object-specific experts into a shared Transformer policy.
RobustDexGrasp~\cite{zhang2025robustdexgrasp} grasps general objects
robustly from single-view perception, and GraspXL~\cite{zhang2024graspxl}
generates grasping motions at scale for several hand models.
DexNDM~\cite{liu2026dexndm} addresses the reality gap in in-hand rotation
using joint-wise learned dynamics. These studies target object
generalization and deployment for one embodiment; we share a grasping
policy when both the arm and hand vary.

\begin{figure*}[t]
    \centering
    \includegraphics[width=\textwidth]{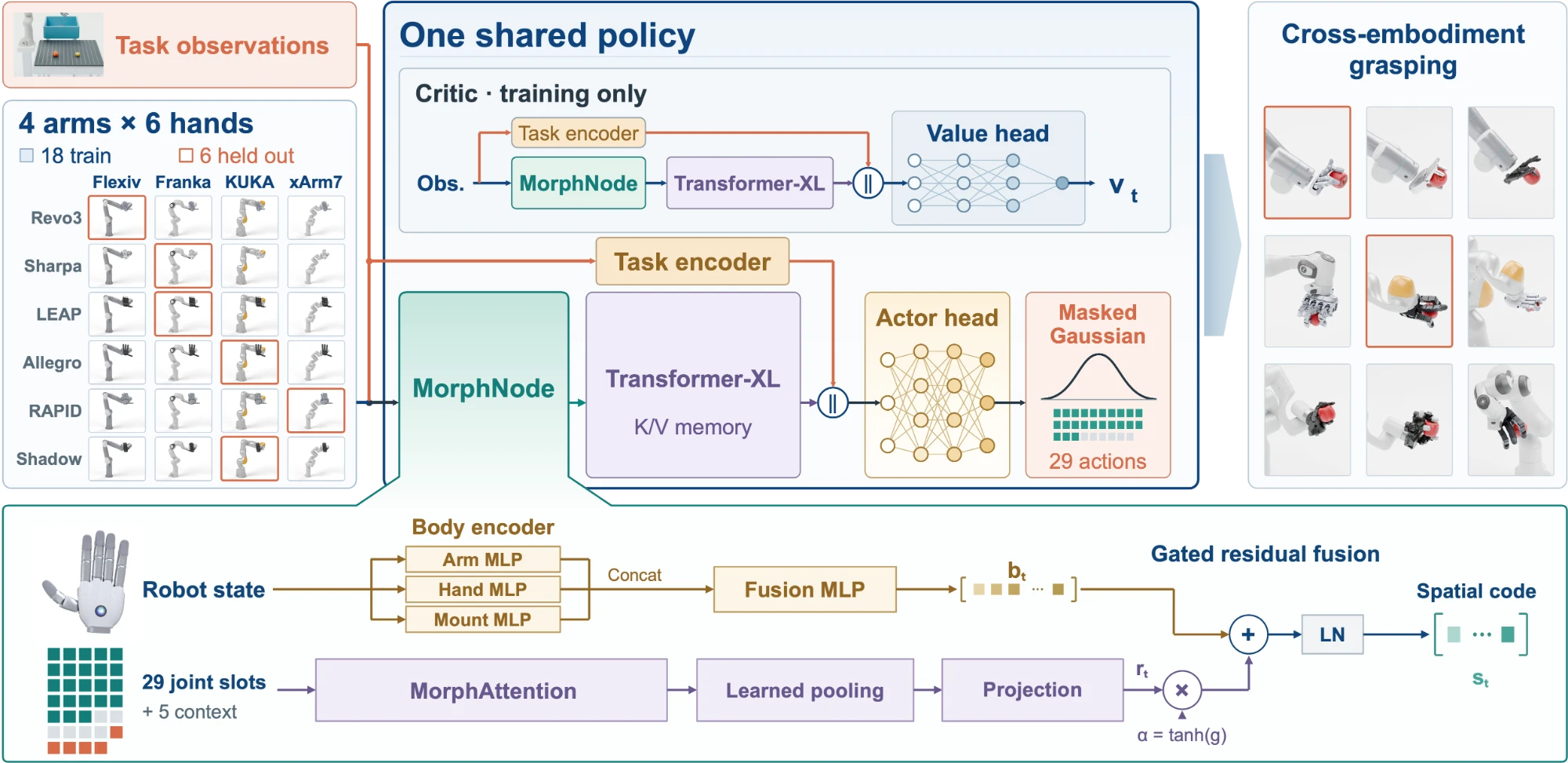}
    \caption{\textbf{DexFormer 2.0 architecture.}
    Top: one shared policy controls all 24 arm--hand combinations (18 for
    training, 6 \heldterm{held out}). MorphNode encodes the canonical joint state,
    Transformer-XL adds a 16-step key/value \memterm{memory}, and a task encoder
    feeds the current task observations directly to the actor head, which
    outputs a masked Gaussian over 29 actions. The critic, used only in
    training, has the same structure with its own weights and \memterm{memory}.
    Bottom: MorphNode. A body encoder (arm, hand, and mount MLPs) gives
    $\mathbf{b}_t$; MorphAttention attends over joint and context tokens
    with topology biases and pools them to $\mathbf{r}_t$; gated fusion
    gives the spatial code
    $\mathbf{s}_t=\operatorname{LN}(\mathbf{b}_t+\alpha\mathbf{r}_t)$.}
    \label{fig:architecture}
\end{figure*}

\paragraph{Generalist policies across embodiments.}
Morphology-agnostic control has a long line of work in simulated
locomotion. Graph and modular policies share parameters across limbs
through message passing~\cite{wang2018nervenet,huang2020smp}, and
Transformer policies replace message passing with attention over joint
tokens, optionally conditioned on morphology~\cite{kurin2021amorpheus,gupta2022metamorph}.
For manipulation, large imitation-learning models pool demonstrations
from many robots~\cite{openx2024,octo2024,doshi2024crossformer}, but
they mostly cover parallel-jaw grippers.
One Policy to Run Them All~\cite{bohlinger2025onepolicy} introduces the
Unified Robot Morphology Architecture (URMA), using morphology-aware
encoding and decoding to share a locomotion policy across diverse legged
robots, later scaled through embodiment
randomization~\cite{bohlinger2025extreme}.
RMA~\cite{kumar2021rma} adapts a locomotion policy to changing dynamics
from a \memterm{memory} of recent states and actions. Dexterous grasping additionally requires coordinated arm
motion and multi-finger contact control.

\paragraph{Cross-embodiment dexterous control.}
CrossDex~\cite{yuan2025crossdex} learns a shared grasping policy using
human-hand eigengrasp actions and hand-specific retargeting.
GET-Zero~\cite{patel2025getzero} incorporates kinematic graph structure
into a Transformer for transfer across hand designs in object rotation.
DexFormer~\cite{zhang2026dexformer} instead conditions on a \memterm{memory} of
past observations to infer morphology and dynamics implicitly.
DexGrasp-Zero~\cite{wu2026dexgraspzero} uses a morphology-aligned graph
representation and physical-property injection for cross-hand grasping.
Building on these lines of research, DexFormer 2.0 studies
\cut{\emph{joint arm-and-hand embodiment variation}: one grasping policy is
shared across the Cartesian product of four arms (Flexiv, Franka, KUKA, and
xArm7) and six hands (Revo3, Sharpa, LEAP, Allegro, RAPID, and Shadow), yielding
$4\times6=24$ arm--hand combinations.}

\section{Method}
\label{sec:method}

\rev{We train one policy that controls different arm--hand combinations in
joint space and generalizes to \heldterm{held-out arm--hand pairs} and to entirely new
hand types, as shown in \cref{fig:architecture}. Each
robot is mapped to shared joint slots (\cref{sec:problem-formulation}),
each described by its joint's physical properties
(\cref{sec:morphology-representation}). MorphNode encodes them into a
spatial code (\cref{sec:morphnode}), a Transformer-XL adds \memterm{memory} of
recent steps alongside a task encoder (\cref{sec:temporal-context}), and
\sharedmark{an actor head} outputs masked joint actions
(\cref{sec:action-decoder}). Training uses PPO on a mixture of
embodiments (\cref{sec:training-objective}). \sharedmark{All network parameters are shared
across embodiments; embodiments differ only in their inputs (joint
descriptors, validity masks, and tree distances), and each environment
keeps its own Transformer-XL \memterm{memory}.}}

\subsection{Problem Formulation}
\label{sec:problem-formulation}
Let $e=(a,h)$ denote an embodiment formed by arm $a\in\mathcal{A}$ and
hand $h\in\mathcal{H}$. We seek one stochastic policy
$\pi_\theta(\mathbf{u}_t\mid\mathbf{o}_{\leq t},\mathbf{m}_e)$ that controls
all embodiments in $\mathcal{A}\times\mathcal{H}$. The observation
$\mathbf{o}_t$ contains the robot joint state, object pose and target
position, fingertip states and contact forces, and a 64-point object
point cloud. The morphology description
$\mathbf{m}_e$ is derived from the robot model and joint limits. The action
$\mathbf{u}_t$ specifies relative joint targets for both the arm and hand.

All embodiments are mapped to a canonical interface with $N_a=7$ arm slots
and $N_h=22$ hand slots. A binary mask
$\mathbf{v}_e\in\{0,1\}^{N_a+N_h}$ marks physically active joints. Padding
slots have zero observations, descriptors, action means, and policy-loss
contributions, so inactive joints affect neither execution nor learning.

\subsection{Joint-Level Morphology Representation}
\label{sec:morphology-representation}
For each canonical joint $i$, we construct a descriptor
$\mathbf{m}_i\in\mathbb{R}^{16}$. It contains the joint axis expressed in its
parent-link frame (3 dimensions), joint attachment position (3), the center
and half-range of its position limits (2), velocity limit (1), a semantic
one-hot role for arm or finger identity (6), and normalized kinematic depth
(1). Unlike a categorical robot identifier, the descriptor exposes local
physical structure that can be shared with \heldterm{held-out pairs}.

Each joint combines its descriptor with six state features: position,
velocity, previous action, and a three-dimensional joint location. This
forms a 22-dimensional feature $\mathbf{x}_{i,t}$, accompanied by the
validity mask. The mount pose, i.e., the relative wrist-to-palm pose,
contributes seven additional values. Kinematic ancestry is derived from the robot model and used to
compute joint-to-joint tree distances.

\subsection{MorphNode Spatial Encoder}
\label{sec:morphnode}
MorphNode combines a body MLP path with a topology-aware
MorphAttention residual. In the body path, masked joint features and
validity bits are flattened separately for the arm and hand and encoded
by an arm encoder, a hand encoder, and a mount encoder for the mount
pose. Each output is adapted to 128 dimensions, and a fusion MLP produces
the body feature $\mathbf{b}_t$. The arm, hand, and mount encoders are
MLPs with layer widths 128--96, 64--96, and 64--64, and the fusion MLP
has two 128-unit layers.

The MorphAttention residual projects the 29 joint features to 128-dimensional
tokens and appends five context tokens: object state, goal state, contacts,
remaining task observations, and mount pose. Two spatial attention layers,
each with four heads, update these 34 tokens. For joint tokens $i,j$, the
attention logits include learned tree-distance and joint-role biases,
in the spirit of structural encodings for graph
Transformers~\cite{ying2021graphormer}:
\begin{equation}
    \ell_{ij}=\frac{\mathbf{q}_i^\top\mathbf{k}_j}{\sqrt{d}}
      + B_{\mathrm{dist}}(D_{ij}) + B_{\mathrm{role}}(R_{ij}).
    \label{eq:topology-attention}
\end{equation}
Here $R_{ij}$ distinguishes same-role, different-finger, and arm--hand
relations. These are soft attention preferences, not hard connectivity
constraints. Context tokens attend globally without an assigned kinematic
location; invalid joint tokens are masked. Learned weighted pooling over
valid tokens gives a 128-dimensional vector. Concatenating the fractions of
valid arm and hand joints and projecting $130\!\rightarrow\!128$ yields
the residual $\mathbf{r}_t$. Residual fusion gives the temporal input
\begin{equation}
    \mathbf{s}_t=\operatorname{LN}(\mathbf{b}_t+\alpha\mathbf{r}_t),
    \qquad \alpha=\tanh(g),
    \label{eq:morphnode-fusion}
\end{equation}
where $g$ is learned and $\alpha$ is initialized to $0.1$.

\subsection{Morphology-Conditioned \memterm{Memory}}
\label{sec:temporal-context}
A three-layer, eight-head Transformer-XL~\cite{dai2019transformerxl} processes the 128-dimensional
spatial features $\mathbf{s}_t$ with \memterm{memory length} 16 and a 512-dimensional
SwiGLU feed-forward hidden layer:
\begin{equation}
    \mathbf{c}^{\mathrm{hist}}_t
    = \operatorname{TXL}(\mathbf{s}_t,
      \mathbf{c}^{\mathrm{mem}}_{t-1}).
    \label{eq:txl-context}
\end{equation}
Each layer maintains a \memterm{key/value memory} independently for every
environment, resetting it at episode boundaries. This \memterm{memory} lets the policy infer interaction dynamics and
embodiment-dependent response that are not fully specified by static geometry.

Current task observations---including object and target state, contacts,
perception, and fingertip states---are normalized with validity masks so that
absent fingers do not bias running statistics. A
task encoder, a $300\!\rightarrow\!192\!\rightarrow\!96$ MLP followed
by a 128-dimensional adapter, maps them to
$\mathbf{c}^{\mathrm{task}}_t\in\mathbb{R}^{128}$. This path
bypasses the temporal module. The context passed to the heads is
\begin{equation}
    \mathbf{c}_t =
    [\mathbf{c}^{\mathrm{hist}}_t;
     \mathbf{c}^{\mathrm{task}}_t]
    \in\mathbb{R}^{256}.
    \label{eq:global-context}
\end{equation}

\subsection{Actor and Value Heads}
\label{sec:action-decoder}
The actor head, a single MLP, maps the 256-dimensional context to the
canonical action means:
\begin{equation}
    \boldsymbol{\mu}_t=\mathbf{v}_e\odot f_{\theta}(\mathbf{c}_t),
    \label{eq:global-head}
\end{equation}
where $f_{\theta}$ has widths
$256\!\rightarrow\!256\!\rightarrow\!256\!\rightarrow\!29$ and is
shared across embodiments. \cut{The mask removes inactive action dimensions.}

The actor is a diagonal Gaussian with learnable, state-independent
standard deviations bounded to $[0.05,1.0]$. Sampled actions are clipped
to $[-5,5]$, scaled by $0.05$ for arm joints and $0.08$ for hand joints,
and applied as relative joint-position targets. The critic uses the same
architecture with a scalar value head, is used only in training, and
shares no weights or \memterm{memory} with the actor.

\begin{figure*}[t]
    \centering
    \begin{subfigure}[b]{0.533\textwidth}
        \centering
        \includegraphics[width=\linewidth]{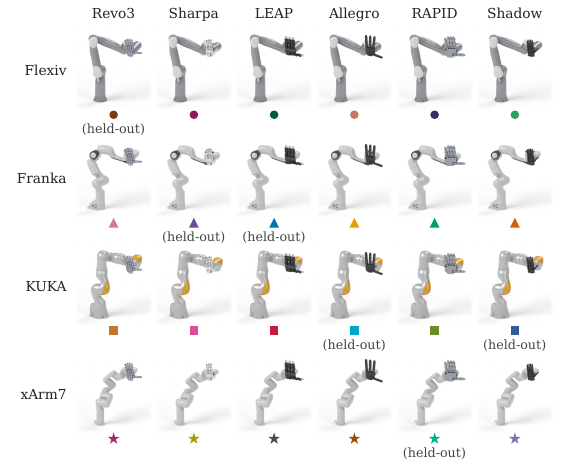}
        \caption{Arm--hand embodiments}
    \end{subfigure}\hfill
    \begin{subfigure}[b]{0.452\textwidth}
        \centering
        \includegraphics[width=\linewidth]{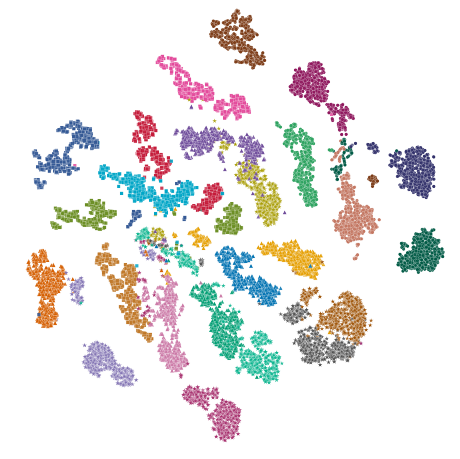}
        \caption{Temporal representations}
    \end{subfigure}
    \caption{Embodiments and temporal representations of the shared policy.
    (a) Four arms and six hands form 24 combinations, shown in static
    explanatory poses. The colored symbols provide the key for (b);
    only the six \heldterm{held-out pairs} are labeled \textnormal{(\heldterm{held-out})}.
    (b) t-SNE of 19,200 actor Transformer-XL outputs (800 episodes per
    embodiment at 1\,s) from the no-fault checkpoint at iteration 5,000.
    Colors identify arm--hand pairings;
    shapes identify arms: circles for Flexiv, triangles for Franka,
    squares for KUKA, and stars for xArm7.}
    \label{fig:embodiment-representations}
    \label{fig:embodiment-grid}
    \label{fig:txl-tsne}
\end{figure*}

\subsection{Training Objective}
\label{sec:training-objective}
We train the shared actor--critic with PPO~\cite{schulman2017ppo} over a mixture of embodiments.
Each rollout contains environments with different arm--hand pairings, so
every update aggregates gradients across morphologies. The policy runs at
60\,Hz in a 240\,Hz simulation with 64-step rollouts. Rewards encourage
fingertip reaching, thumb-and-finger contact, and target tracking once
in contact, with a success bonus and action penalties. A curriculum
raises gravity from zero to full, and physical parameters are
randomized (Supp.~Sec.~A).
\cut{Invalid canonical slots stay masked in observation normalization, action
sampling, and the policy loss.}

\begin{figure*}[t]
    \centering
    \begin{subfigure}[b]{0.44\textwidth}
        \centering
        \includegraphics[width=\linewidth]{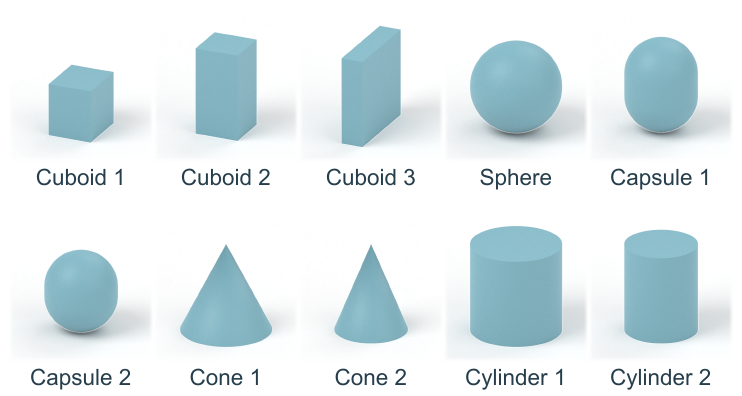}
        \caption{Seen objects: primitives training}
    \end{subfigure}\hspace{0.02\textwidth}
    \begin{subfigure}[b]{0.44\textwidth}
        \centering
        \includegraphics[width=\linewidth]{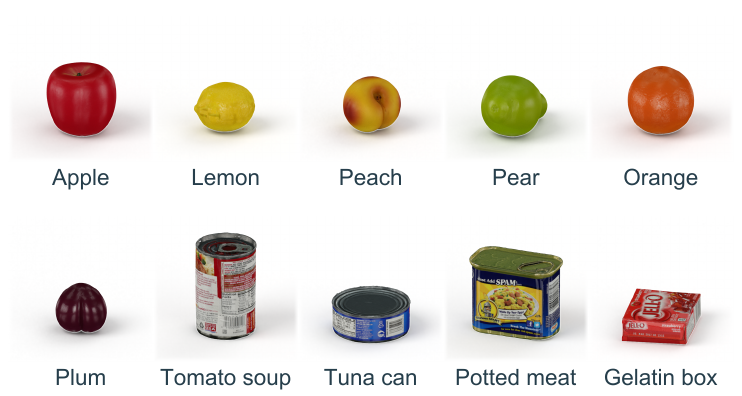}
        \caption{Unseen objects: YCB test}
    \end{subfigure}
    \caption{Training and evaluation objects. Left: the ten seen
    primitive configurations from five shape families. Right: ten
    unseen YCB objects~\cite{calli2015ycb}. All objects share one
    metric scale and camera view.}
    \label{fig:object-sets}
\end{figure*}

\begin{figure}[t]
    \centering
    \includegraphics[width=\linewidth]{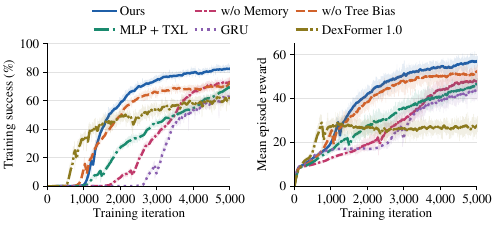}
    \caption{Training success and mean episode reward on the 18 training
    combinations \seedmark{(trailing 50-iteration mean)}.}
    \label{fig:training-curves}
\end{figure}

\section{Experiments}
\label{sec:experiments}

\subsection{Experimental Setup}
\label{sec:experimental-setup}
\paragraph{Embodiments.}
Four arms (Flexiv, Franka, KUKA, and xArm7) and six hands (Revo3, Sharpa,
LEAP, Allegro, RAPID, and Shadow) form 24 embodiments
(\cref{fig:embodiment-representations}a). A fixed split uses 18
combinations for training and holds out Franka--LEAP, Franka--Sharpa,
KUKA--Allegro, KUKA--Shadow, xArm7--RAPID, and Flexiv--Revo3.

\paragraph{Objects.}
Training uses ten primitive configurations from five shape families at
fixed nominal size: three cuboids, one sphere, two capsules, two cones,
and two cylinders. For object generalization, we evaluate ten YCB
objects~\cite{calli2015ycb} excluded from training: apple, lemon, peach,
pear, orange, plum, tomato soup can, tuna can, potted meat can, and
gelatin box (\cref{fig:object-sets}).

\paragraph{Metric.}
\label{sec:evaluation-metric}
Our primary metric is the stable success rate (SR), the percentage of
evaluation episodes that succeed. An episode succeeds if it completes all
240 control steps, keeps the object position error below 5\,cm throughout
the final 30 steps (0.5\,s), and has no non-timeout termination. All
episodes remain in the denominator, and split-level results average the
per-embodiment rates equally.

\paragraph{Protocol.}
All policies are trained with PPO~\cite{schulman2017ppo} in Isaac
Lab~\cite{mittal2025isaaclab} and evaluated at iteration 5,000 with frozen
weights and normalization statistics, deterministic actions, no
observation noise, and the final curriculum difficulty. \seedmark{\rev{On seen and unseen objects, each embodiment and condition uses
1,000 episodes. Simulation results are reported over five random seeds.} \rev{Each episode places the object at a random position within
$\pm$0.2\,m of its default position on the table, with a random
orientation, and samples a random target position; all policies use the
same saved scenes.}} Supplementary Section~A gives the remaining details.

\subsection{Comparison with Baselines}
\label{sec:baseline-comparison}

\begin{table}[t]
    \centering
    \caption{Stable SR (\%) $\uparrow$ across object and arm--hand combination splits. Seen objects use the training-primitive mixture; unseen objects average ten YCB objects. \heldterm{Held-out arm--hand pairs} combine arm and hand types that both appear in training. \seedmark{Mean $\pm$ sample standard deviation over five seeds.}}
    \label{tab:baseline-comparison}
    \footnotesize
    \setlength{\tabcolsep}{3pt}
    \begin{tabular*}{\columnwidth}{@{\extracolsep{\fill}}llcc@{}}
        \toprule
        Policy & \shortstack{Arm--hand\\pair} & \shortstack{Seen\\objects} & \shortstack{Unseen\\objects} \\
        \midrule
        \multirow{2}{*}{GRU} & Seen (18) & $64.82 \pm 0.41$ & $65.61 \pm 0.93$ \\
         & \heldterm{Held-out} (6) & $31.28 \pm 0.94$ & $33.18 \pm 1.89$ \\
        \midrule
        \multirow{2}{*}{DexFormer 1.0} & Seen (18) & $66.02 \pm 0.91$ & $70.65 \pm 0.49$ \\
         & \heldterm{Held-out} (6) & $14.15 \pm 0.58$ & $11.53 \pm 0.60$ \\
        \midrule
        \multirow{2}{*}{Ours} & Seen (18) & $\mathbf{89.01 \pm 0.56}$ & $\mathbf{89.01 \pm 0.31}$ \\
         & \heldterm{Held-out} (6) & $\mathbf{51.18 \pm 1.72}$ & $\mathbf{48.85 \pm 1.94}$ \\
        \bottomrule
    \end{tabular*}
\end{table}

\begin{figure*}[t]
    \centering
    \includegraphics[width=\textwidth]{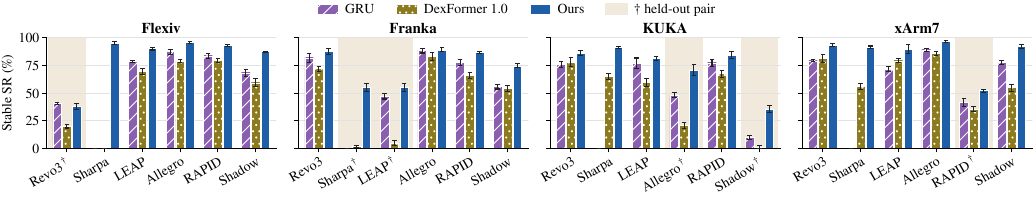}
    \caption{Stable SR (\%) for all 24 arm--hand combinations on seen
    objects (1,000 episodes each). \seedmark{Error bars: sample standard deviation
    over five seeds.} Shaded groups marked $\dagger$ are the six \heldterm{held-out pairs}. Exact values: Supp.~Table~7.}
    \label{fig:embodiment-matrix}
\end{figure*}

We compare DexFormer 2.0 (Ours) with two baselines: a
GRU~\cite{cho2014gru} policy with MLP encoders, and
DexFormer 1.0~\cite{zhang2026dexformer}, which conditions on a \memterm{memory} of
past observations without explicit morphology descriptors.
\rev{CrossDex~\cite{yuan2025crossdex} (fixed arm), GET-Zero~\cite{patel2025getzero}
(no arm), and DexGrasp-Zero~\cite{wu2026dexgraspzero} (Cartesian wrist) all
need per-hand action mappings or experts and do not fit our setting
(Supp.~Sec.~B).}

\Cref{tab:baseline-comparison} crosses object and embodiment
generalization, and \cref{fig:embodiment-matrix} gives per-embodiment
rates. On the seen-object mixture, Ours reaches 89.01\% on seen combinations and
51.18\% on \heldterm{held-out pairs}, compared with 64.82\%/31.28\% for GRU
and 66.02\%/14.15\% for DexFormer 1.0.
\cut{On unseen YCB objects, Ours achieves 89.01\%/48.85\% on training/\heldterm{held-out} pairs, versus 65.61\%/33.18\% for GRU and 70.65\%/11.53\%
for DexFormer 1.0.} When the object is unseen and the pair \heldterm{held out},
Ours exceeds the stronger baseline, GRU, by 15.67 percentage points.
The training curves show the same ranking
(\cref{fig:training-curves}, Supp.~Sec.~G).
A t-SNE~\cite{vandermaaten2008tsne} projection of the \memterm{memory} features
(\cref{fig:embodiment-representations}b, Supp.~Sec.~D)
is organized by embodiment, including the \heldterm{held-out pairs}.

\subsection{Ablation Study}
\label{sec:ablation}

\begin{table}[t]
    \centering
    \caption{Network ablations (stable SR, \%). \memterm{Mem.}: 16-step Transformer-XL \memterm{memory}; Attn.: MorphAttention residual; Tree: kinematic tree-distance bias, inapplicable (--) without the residual.}
    \label{tab:ablation}
    \footnotesize
    \setlength{\tabcolsep}{2pt}
    \begin{tabular*}{\columnwidth}{@{\extracolsep{\fill}}lccccc@{}}
        \toprule
        Policy & \memterm{Mem.} & Attn. & Tree & Seen (18) & \heldterm{Held-out} (6) \\
        \midrule
        Full (Ours) & $\checkmark$ & $\checkmark$ & $\checkmark$ & $\mathbf{89.01 \pm 0.56}$ & $\mathbf{51.18 \pm 1.72}$ \\
        w/o \memterm{Memory} &  & $\checkmark$ & $\checkmark$ & $79.14 \pm 1.07$ & $24.45 \pm 0.41$ \\
        w/o Tree Bias & $\checkmark$ & $\checkmark$ &  & $78.43 \pm 0.39$ & $24.15 \pm 1.31$ \\
        MLP + TXL & $\checkmark$ &  & -- & $73.69 \pm 0.54$ & $18.90 \pm 0.72$ \\
        \bottomrule
    \end{tabular*}
\end{table}

\Cref{tab:ablation} evaluates the spatial and temporal components.
The \memterm{memory} ablation sets the Transformer-XL cache length to zero while
retaining the temporal network and current observations. The tree-bias
ablation removes the kinematic tree-distance bias but keeps joint-role
biases and morphology descriptors. MLP + TXL removes the MorphAttention
residual while retaining the body MLP path, task encoder,
Transformer-XL, and heads.

Relative to Ours, removing \memterm{memory} reduces \heldterm{held-out SR} by 26.73 percentage
points; removing tree-distance bias reduces it by 27.03 points.
\cut{MLP + TXL reaches 18.90\% \heldterm{held-out SR}, compared with 51.18\% for Ours.}

\subsection{Effect of Training Embodiment Coverage}
\label{sec:training-coverage}

\begin{table*}[t]
    \centering
    \caption{Training arm--hand combination coverage at a matched interaction budget. The training sets are nested: $\mathcal{T}_6\subset\mathcal{T}_{12}\subset\mathcal{T}_{18}$. Seen and \heldterm{Held-out} use each policy's own split and so differ in composition across rows. \rev{Common \heldterm{held-out} (6) and All (24)} use identical evaluation sets.}
    \label{tab:training-coverage}
    \small
    \setlength{\tabcolsep}{5pt}
    \begin{tabular*}{\textwidth}{@{\extracolsep{\fill}}cccccc@{}}
        \toprule
        \multicolumn{2}{c}{Arm--hand combinations} & \multicolumn{4}{c}{Stable SR (\%) $\uparrow$} \\
        \cmidrule(lr){1-2}\cmidrule(lr){3-6}
        Seen (train) & \rev{\heldterm{Held-out} (zero-shot)} & Seen & \heldterm{Held-out} &  \rev{Common \heldterm{held-out} (6)} & \rev{All (24)} \\
        \midrule
         6 & 18 & $56.58 \pm 0.58$ & $0.08 \pm 0.05$ & $0.07 \pm 0.07$ & $14.21 \pm 0.18$ \\
        12 & 12 & $61.07 \pm 0.31$ & $3.37 \pm 0.22$ & $5.65 \pm 0.41$ & $32.22 \pm 0.24$ \\
        18 &  6 & $89.01 \pm 0.56$ & $51.18 \pm 1.72$ & $\mathbf{51.18 \pm 1.72}$ & $\mathbf{79.55 \pm 0.73}$ \\
        \bottomrule
    \end{tabular*}
\end{table*}

\begin{table*}[t]
    \centering
    \caption{\textbf{Transfer to unseen hand and gripper types.} Stable SR (\%) on seen objects with the frozen policy. \cut{All hand and gripper types are absent from training and paired with seen arm types. The policy directly controls the dexterous hands.} The grippers are driven by the policy's own actions through a fixed virtual-hand retargeting map (Supp.~Sec.~C), in which the policy observes a virtual LEAP hand rather than the gripper.}
    \label{tab:unseen-hand-transfer}
    \footnotesize
    \setlength{\tabcolsep}{2.5pt}
    \begin{tabular*}{\textwidth}{@{\extracolsep{\fill}}lcccccccc@{}}
        \toprule
        & \multicolumn{4}{c}{Unseen hand types (direct policy control)} & \multicolumn{4}{c}{Unseen gripper types (virtual-hand retargeting)} \\
        \cmidrule(lr){2-5}\cmidrule(lr){6-9}
        Seen arm & \rev{Revo2} & Wuji 1 & \tocheck{Tesollo DG-5F-M} & XHAND1 & Franka Hand & Robotiq 2F-85 & xArm G1 & UMI WSG-50 \\
        \midrule
        Flexiv & \rev{\revtwo{$\mathbf{26.60 \pm 2.22}$}} & $\mathbf{41.20 \pm 2.80}$ & $12.70 \pm 3.88$ & $26.50 \pm 4.47$ & $14.70 \pm 0.91$ & $7.00 \pm 0.61$ & $5.90 \pm 0.82$ & $3.70 \pm 1.35$ \\
        Franka & \rev{\revtwo{$4.60 \pm 1.08$}} & $10.00 \pm 2.03$ & $7.00 \pm 1.27$ & $6.10 \pm 1.29$ & $18.70 \pm 1.30$ & $\mathbf{16.40 \pm 3.23}$ & $\mathbf{20.20 \pm 2.05}$ & $\mathbf{15.10 \pm 1.78}$ \\
        KUKA & \rev{\revtwo{$14.40 \pm 2.22$}} & $24.10 \pm 1.24$ & $6.70 \pm 1.35$ & $6.80 \pm 1.44$ & $14.30 \pm 1.72$ & $8.10 \pm 1.19$ & $6.10 \pm 0.74$ & $3.60 \pm 1.56$ \\
        xArm7 & \rev{\revtwo{$10.70 \pm 1.04$}} & $33.40 \pm 2.48$ & $\mathbf{20.70 \pm 2.08}$ & $\mathbf{27.70 \pm 2.84}$ & $\mathbf{18.80 \pm 3.21}$ & $15.00 \pm 1.41$ & $13.10 \pm 1.85$ & $7.80 \pm 2.11$ \\
        \midrule
        Mean & \rev{\revtwo{$14.07 \pm 0.91$}} & $27.18 \pm 1.32$ & $11.78 \pm 1.44$ & $16.78 \pm 1.87$ & $16.62 \pm 0.83$ & $11.62 \pm 1.00$ & $11.32 \pm 0.58$ & $7.55 \pm 0.90$ \\
        \bottomrule
    \end{tabular*}
\end{table*}

We vary the number of training arm--hand combinations from 6 to 12 to 18
while keeping all four arm types and all six hand types in every training
set. The architecture, optimizer, rollout length, environment count, and
checkpoint iteration are unchanged, so the interaction budget is matched
and fewer training embodiments receive more samples each.

\Cref{tab:training-coverage} reports transfer to the same six \heldterm{held-out pairs} in every run, which fixes the evaluation set while
training coverage changes. \cut{Across all 24 embodiments, SR rises from 14.21\% to 32.22\% to 79.55\%.}
On the common \heldterm{held-out set}, SR increases from 0.07\% to 5.65\% to 51.18\%
as training coverage grows from 6 to 12 to 18 combinations. The 18-combination
policy gains 45.53 percentage points over the 12-combination policy on this
fixed test set. \cut{All arm and hand types are present even in the smallest
training set, so the gain comes from pairing coverage.}

\subsection{Transfer to Unseen Hand and Gripper Types}
\label{sec:unseen-hand-transfer}

The \heldterm{held-out pairs} in \cref{fig:embodiment-matrix} use arm and hand
types encountered during training. To test transfer to new hands, we pair
each of the four training arms with \rev{four} previously unseen right hands.
Each hand fills the canonical slots by joint role
(Supp.~Sec.~A), supplies its morphology descriptors to the
same frozen policy, and is evaluated on the ten seen objects.

Wuji 1 has the highest SR on all four arms (\cref{tab:unseen-hand-transfer}),
with a mean of 27.18\%.
\tocheck{Tesollo achieves 11.78\% on average and performs best on xArm7
(20.70\%).} \cut{XHAND1 achieves 16.78\% on average, with 26.50\% on Flexiv
and 27.70\% on xArm7 but below 7\% on Franka and KUKA.} \rev{Revo2 achieves \revtwo{14.07\%} on average and performs best on Flexiv (\revtwo{26.60\%}).} Across the
\rev{four} hands, the policy achieves \rev{\revtwo{17.45\%}} mean SR.

\paragraph{Parallel grippers.}
\rev{The gripper columns of} \cref{tab:unseen-hand-transfer} ask whether the same frozen
policy can drive a gripper with a single opening degree of freedom.
Mapping the thumb and index commands directly to the jaws almost never
yields a stable grasp. We therefore let the policy observe and
command a virtual LEAP hand~\cite{shaw2023leap} attached to the real
flange. Its arm actions are executed unchanged, and the jaw opening
follows the distance between the virtual thumb and index fingertips
(Supp.~Sec.~C). With this fixed map, mean SR ranges from
7.55\% for UMI to 16.62\% for the Franka Hand. Because the policy
observes a LEAP hand here, this shows that its actions can be retargeted
to a gripper, not zero-shot transfer to gripper morphology.

\begin{figure*}[t]
    \centering
    \includegraphics[width=\textwidth]{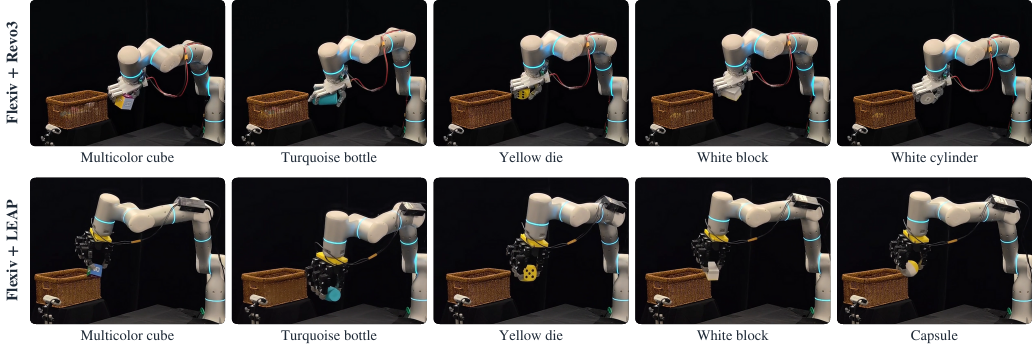}
    \caption{\textbf{Real-world grasping tasks.} A Flexiv arm with a Revo3
    hand (top) and a LEAP hand (bottom) grasps tabletop objects and lifts
    them toward a basket. Each image shows one task during the lift; the
    first four columns show the same object for both hands.}
    \label{fig:real-world-tasks}
\end{figure*}

\subsection{Robustness to Hand-Joint Faults}
\label{sec:sim-fault-injection}

\begin{table}[t]
    \centering
    \caption{\rev{Stable SR (\%) $\uparrow$ under hand-joint target-hold faults. Initial: one joint target held at zero from episode start; Mid: its measured angle held after 1\,s. 1,000 episodes per embodiment and condition.}}
    \label{tab:sim-fault-injection}
    \footnotesize
    \setlength{\tabcolsep}{3pt}
    \begin{tabular*}{\columnwidth}{@{\extracolsep{\fill}}lcccccc@{}}
        \toprule
        & \multicolumn{3}{c}{Seen (18)} & \multicolumn{3}{c}{\heldterm{Held-out} (6)} \\
        \cmidrule(lr){2-4}\cmidrule(lr){5-7}
        Policy & Normal & Initial & Mid & Normal & Initial & Mid \\
        \midrule
        GRU & 64.82 & 59.51 & 62.95 & 31.28 & 28.53 & 30.77 \\
        DexFormer 1.0 & 66.02 & 56.61 & 62.43 & 14.15 & 11.70 & 12.25 \\
        Ours & \textbf{89.01} & \textbf{81.74} & \textbf{86.40} & \textbf{51.18} & \textbf{45.53} & \textbf{49.63} \\
        \bottomrule
    \end{tabular*}
\end{table}

We test hand-joint target-hold faults with the frozen checkpoints \rev{(\cref{tab:sim-fault-injection})}. An initial fault holds one hand joint's
execution target at zero from episode start; a mid-episode fault holds
the joint's measured angle as its target after 1\,s. \rev{Ours loses at most 7.27 percentage points on seen combinations and 5.65 on \heldterm{held-out pairs}, and stays above both baselines in every condition.}
Training with the same faults reduces the drop under initial holds from
7.27 to 5.12 percentage points, but lowers SR on \heldterm{held-out pairs} to
34.75\% (Supp.~Sec.~F). \rev{We therefore train Ours without fault
injection.}

\subsection{Real-World Experiments}
\label{sec:real-world-main}

We deploy on a physical Flexiv arm with two hands, Revo3 and LEAP
(\cref{fig:real-world-tasks}). The simulation policy observes object
state that is unavailable on the robot, so \tocheck{we distill it into a
student policy} that observes depth from two stereo cameras and joint
feedback. The student is trained in simulation with online
DAgger~\cite{ross2011dagger} under randomized camera poses, depth noise,
object size, and initial states (Supp.~Sec.~H), \rev{and
is then deployed on the robot zero-shot, without real-world fine-tuning.} Once the
predicted object position stays within 5\,cm of the target for four
consecutive new frames, a scripted routine holds the grasp, lifts and
carries the object to a basket, and releases it. A trial succeeds if
the object ends up in the basket.
\rev{Each hand grasps each object five times
(\cref{tab:real-world-main}): Revo3 succeeds in 17 of 25 trials (68\%)
and LEAP in 23 of 25 (92\%).}
\rev{Flexiv + Revo3 is one of the six \heldterm{held-out pairs} of the
simulation policy, whereas Flexiv + LEAP is a training pair. As in
simulation, the \heldterm{held-out pair} succeeds less often than the training pair.
Real-world and simulation success rates are not directly
comparable, because the two evaluations use different objects, scene
distributions, and success criteria.}

\begin{table}[!t]
    \centering
    \caption{Real-world grasping successes out of \rev{5} trials per object.
    Obj.~1--5 are the objects in columns 1--5 of
    \cref{fig:real-world-tasks}. \rev{\heldterm{Held-out}: excluded from the training of
    the simulation policy; training: a training pair.}}
    \label{tab:real-world-main}
    \footnotesize
    \setlength{\tabcolsep}{2pt}
    \renewcommand{\arraystretch}{1.25}
    \begin{tabular*}{\columnwidth}{@{\extracolsep{\fill}}l*{6}{c}@{}}
        \toprule
        Embodiment & Obj.~1 & Obj.~2 & Obj.~3 & Obj.~4 & Obj.~5 & Total \\
        \midrule
        Flexiv + Revo3 \rev{(\heldterm{held-out})} & \rev{4/5} & \rev{2/5} & \rev{5/5} & \rev{3/5} & \rev{3/5} & \rev{17/25} \\
        Flexiv + LEAP \rev{(training)} & \rev{5/5} & \rev{4/5} & \rev{5/5} & \rev{5/5} & \rev{4/5} & \rev{23/25} \\
        \bottomrule
    \end{tabular*}
\end{table}

\section{Conclusion}
\label{sec:conclusion}
We presented DexFormer 2.0, a single reinforcement learning policy that
controls heterogeneous arm--hand systems through a canonical joint
interface, per-joint morphology descriptors, topology-biased attention,
and a short Transformer-XL \memterm{memory}. Trained on 18 of 24 arm--hand
combinations, it reaches 89.01\% stable success on seen and 51.18\% on \heldterm{held-out arm--hand pairs}, and 48.85\% when the objects are also unseen.
\memterm{Memory}, the kinematic tree bias, and pairing coverage are each
necessary. Without fine-tuning, the policy also grasps
with \rev{four} unseen hand types, tolerates unseen joint faults,
and drives parallel grippers through retargeting. On a real Flexiv arm,
distilled depth-based students \rev{succeed in 68\% (Revo3, \heldterm{held-out} pair) and 92\% (LEAP) of trials without real-world fine-tuning.}

{\small
\bibliographystyle{ieeenat_fullname}
\bibliography{references}
}

\end{document}